**Faculty of Natural and Mathematical Sciences**
*Department of Informatics*

King's College London,
Strand Campus, London,
United Kingdom

**KING'S College LONDON**

7CCSMPRJ
Individual Project Submission 2019/20

Name: Jun Wang
Student Number: 1913824
Degree Programme: Artificial Intelligence
Project Title: Two-phase weakly supervised object detection with pseudo ground truth mining
Supervisor: Miaojing Shi
Word count:11163

---

RELEASE OF PROJECT

Following the submission of your project, the Department would like to make it publicly available via the library electronic resources. You will retain copyright of the project.

---

√ I **agree** to the release of my project

☐ I **do not** agree to the release of my project

**Signature:** 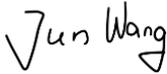

Date: 10, August, 2020

Department of Informatics
King's College London
United Kingdom

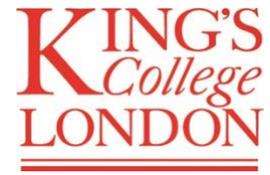

7CCSMPRJ MSc Project

# TWO-PHASE WEAKLY SUPERVISED OBJECT DETECTION WITH PSEUDO GROUND TRUTH MINING

**Name: Jun Wang**
**Degree Programme: Artificial Intelligence**
**Supervisor's Name: Miaojing Shi**

**This dissertation is submitted for the degree of MSc in Artificial Intelligence**

# ACKNOWLEDGEMENT

I would like to express my sincere gratitude to my supervisor Dr. Miaojing Shi for providing me the opportunity to work on this project which is of great interest to me. Without his expert guidance, interpretation and patience, it is not possible to complete this project.
In addition, I thank my families and friends for supporting and encouraging me during this period of my life.

# ABSTRACT

Weakly Supervised Object Detection (WSOD), aiming to train detectors with only image-level dataset, has arisen increasing attention for researchers. In this project, we focus on two-phase WSOD architecture which integrates a powerful detector with a pure WSOD model. We explore the effectiveness of some representative detectors utilized as the second-phase detector in two-phase WSOD and propose a two-phase WSOD architecture. In addition, we present a strategy to establish the pseudo ground truth (PGT) used to train the second-phase detector. Unlike previous works that regard top one bounding boxes as PGT, we consider more bounding boxes to establish the PGT annotations. This alleviates the insufficient learning problem caused by the low recall of PGT. We also propose some strategies to refine the PGT during the training of the second detector. Our strategies suspend the training in specific epoch, then refine the PGT by the outputs of the second-phase detector. After that, the algorithm continues the training with the same gradients and weights as those before suspending. Elaborate experiments are conducted on the PASCAL VOC 2007 dataset to verify the effectiveness of our methods. As results demonstrate, our two-phase architecture improves the mAP from 49.17% to 53.21% compared with the single PCL model. Additionally, the best PGT generation strategy obtains a 0.7% mAP increment. Our best refinement strategy boosts the performance by 1.74% mAP. The best results adopting all of our methods achieve 55.231% mAP which is the state-of-the-art performance.

# CONTENTS

# LIST OF TABLES

# LIST OF FIGURES

# 1 INTRODUCTION

## 1.1 Problem Statement and Motivation

Object detection aims to detect and localize different semantic instances in images or videos with bounding boxes and obtain their categories. It is a significant sub-field of Computer Vision and has been extensively applied to many fields such as autonomous vehicles, security monitoring, and robotics. The decisions or results from the detector are highly correlated to human safety in applications. Therefore, the precision of object detection is of great significance. However, there are complex challenges since one image may contain different classes of objects and these instances may occur in different parts of images or videos. Fortunately, great success can be achieved in object detection because of the development of Deep Learning techniques and large-scale datasets with instance-level annotations. Convolutional Neural Network (CNN) is the most successful deep learning network in Computer Vision due to its strong feature extraction ability. It has proven its performance in many tasks such as image recognition, image caption and object detection. In object detection, CNN is utilized to generate proposal features from the images or region proposals, and then these features are sent to following layers (e.g. fully connected layers) to perform regression of bounding boxes and classification respectively. The general structure of a Region-based CNN network in object detection is shown in Fig. 1.

As stated before, one of the deterministic factors for the great achievement of deep learning methods in object detection tasks are the large-scale datasets with instance-level annotations. With these datasets, the number of objects, their classes, and their bounding boxes are available during the training of detector. However, it is very costly and time-consuming to build such an instance-level dataset that relies on manual laboring, while image-level labels can easily be obtained from search engines such as Google. Deploying an object detection application with only image-level dataset can curtail the cost of developers. Therefore, it is worth paying more attention to the development of object detection with limited supervision. The purpose of this project is to advance the development of weakly supervised object detection (WSOD) that trains detectors with only image-level annotations. Due to the superior advantages and power of CNN mentioned above, our work adopts CNN based models to explore strategies advancing weakly supervised object detection.

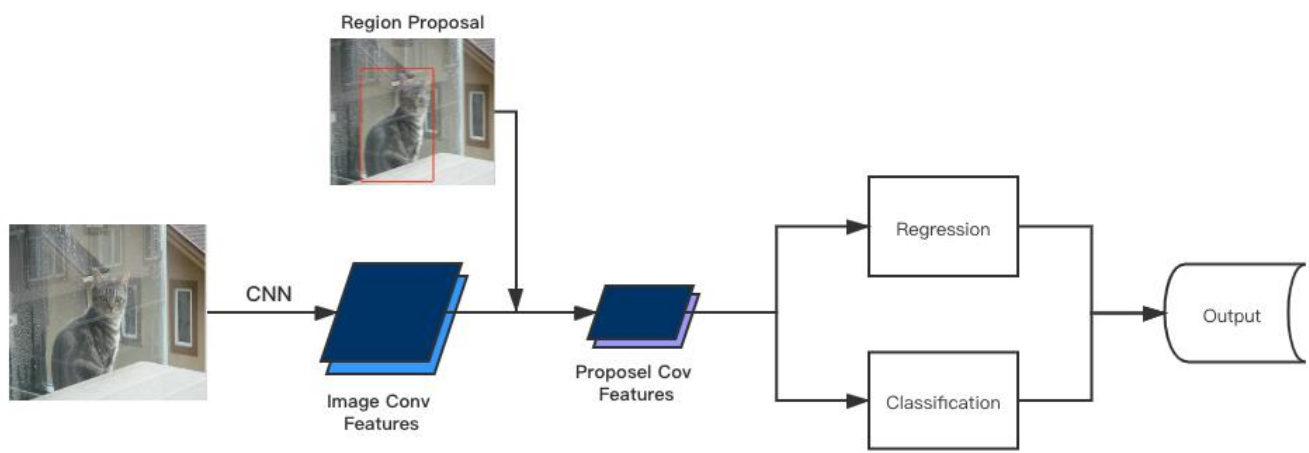

Figure 1. The general structure of Region-based CNN used in object detection. CNN extracts features from images. Algorithms are then utilized to generate proposal features accordning to the image features and region proposals. Next, classification and regression are performed on these proposal feastures to obtain the category and bounding boxes of the potential instance.

## 1.2 Background

Before introducing the aims of our work, some background should be explained first. Many previous works exploit the Multiple Instance Learning (MIL) [1, 2, 3, 4, 5, 6, 7, 8] strategy for WSOD. Moreover, architectures combining the MIL and CNN demonstrate better performance due to the significant feature extraction ability of CNN compared with traditional hand-designed features. Research has shown that an end-to-end training or its variant can further improve the results for WSOD.

Recent powerful WSOD methods normally follow a two-phase learning approach. A MIL [9] detector combined with the CNN feature extractor are trained with image-level labels in the first phase. Outputs of the first-phase model are then taken as pseudo ground truth (PGT) to fine-tune a second-phase detector (e.g. Fast R-CNN [10]). The general structure of this two-phase architecture is shown in Fig. 2. Experiment results have shown that some of these methods achieve competitive performance.

However, we observe that although some research has utilized different single WSOD models as the first-phase detector, the second-phase CNN seems to be the same (Fast R-CNN). Recently, some more powerful CNN detectors have been proposed, such as Faster RCNN [11], Cascade Mask RCNN [12] and EfficientDet [13] Networks. Hence, it is worth investigating the performance of these detectors when implementing them as the second-phase model for the two-phase WSOD architecture. In addition, most of two-phase works regard the top one bounding boxes as the PGT, but the top one bounding boxes may only contain a part of instances or objects, and thus, may confuse the detectors. Further mining methods are required to deal with this problem.

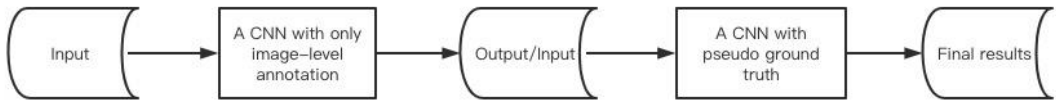

Figure 2. The general architecture of a two-phase training approach. The inputs of the first CNN are images and their region proposals generated by some traditional algorithms such as Selective Search or by a neural network. Outputs generated by the first CNN are regarded as the PGT and sent to the second CNN to fine-tune the model.

## 1.3 Aims and Objectives

The main aim of this work is to advance the development of WSOD.
To summarise, we have the following four objectives:
• To integrate a weakly supervised object detector as the first-phase detector with two or more representative detectors as second-phase networks and test their performance.
• To adopt different strategies for establishing the PGT from the bounding boxes produced by the first-phase WSOD model and test their performance.
• To refine the PGT during the training of the second-detector.
• To analyse the difference among the combinations of networks and strategies, and offer our explanations.
The PACAL VOC 2007 dataset is used to verify the effectiveness of our methods.

## 1.4 Contribution

WSOD is a challenging problem and a certain gap of performance can be seen compared with fully supervised networks. To advance the development of WSOD, we make three methodological improvements. (1) We exploit Fast RCNN and Faster-RCNN with backbone network ResNet [14] to improve the results from a single WSOD detector. (2) Instead of using top one bounding boxes, we regard top $k$ bounding boxes as the PGT to train the second-phase models. (3) During the training of the second-phase detectors, we refine the PGT by the outputs of the second-phase detector.

This work mainly has four contributions: (1) We propose a two-phase WSOD architecture named PCL-Faster RCNN. This model improves the mAP by 4% compared with the single PCL [15] model. (2) We present a strategy to create the PGT. This strategy utilizes more bounding boxes to generate PGT. We then use the PGT to train the second-phase detectors. This method further improves the mAP from 53.20% to 53.98%, which is a competitive performance. (3) We develop some strategies to refine the PGT during the training of the second-phase detector. The best results obtain an absolute 1.74% mAP increment. The best results utilizing all our proposed methods reach 55.231% mAP on the VOC 2007 dataset. (4) We conduct many elaborate experiments with different combinations of PGT generation strategies, network architectures, and their backbone networks on the PASCAL VOC2007 dataset, and offer our results and explanations. These statistics may have implications for or give inspiration to researchers who intend to further advance this field about how to establish PGT, how to select the backbone network, or how to integrate different strategies and components.

# 2 BACKGROUND THEORIES

This section introduces the theories supporting our work and some related works. Section 2.1 and 2.2 describe two representative and well-known object detection networks which are utilized as the second-phase detector in our work. Section 2.3 and 2.4 present some works and techniques to solve WSOD problems. The remaining section describes the state-of-the-art single WSOD model PCL.

## 2.1 Fast RCNN
Proposed by Girshick [10], Fast RCNN is an end-to-end fully supervised object detection model and copes with the slow training speed problem in R-CNN [16]. The structure of Fast RCNN is demonstrated in Fig. 3 [10]. When given an image, the same as R-CNN, select search algorithm is utilized first to generate region proposals. There is a big difference in the next step between Faster RCNN and R-CNN. For Fast RCNN, this image is then sent to a CNN to perform feature extraction. Next, proposal features are generated via mapping image features according to the coordinates of region proposals. For R-CNN, it utilizes CNN on each region proposal to generate proposal features. This means that R-CNN performs feature extraction thousands of times in one image because there are almost 2K region proposals for an image, while Fast RCNN only performs once, and thus, Faster RCNN is more efficient. After that, these proposal features are sent to a ROI pooling layer to generate fixed-size proposal features which are then mapped to feature vectors via a fully connected layer. The last step is classification and regression according to the feature vectors. Fast RCNN exploits two fully connected (FC) layers to perform classification and regression respectively, while R-CNN implements this by some Support Vector Machines (SVM). Compared with fully connected layers, training multiple SVMs separately is complex and time-consuming.

This structure is trained end-to-end with a multi-task loss $L$. The loss function $L$ comprises classification loss $L_{cls}$ and bounding-box regression loss $L_{loc}$ as shown in Equation (1) to (5) [10].

$$L(p, u, t^u, v) = L_{cls}(p, u) + \lambda [u \geq 1] L_{loc}(t^u, v) \quad (1)$$

in which

$$L_{cls}(p, u) = -\log p_u \quad (2)$$

$$[u \geq 1] = \begin{cases} 1 & if\ u \geq 1 \\ 0 & otherwise \end{cases} \quad (3)$$

$$L_{loc}(t^u, v) = \sum_{i \in \{x,y,w,h\}} smooth_{L_1}(t_i^u - v_i) \quad (4)$$

in which

$$smooth_{L_1}(x) = \begin{cases} 0.5x^2 & if\ |x| < 1 \\ |x| - 0.5 & otherwise, \end{cases} \quad (5)$$

Where $p$ is the distribution probability over all classes, $u$ is the ground truth category, $t^u$ is the regression offsets of predicted bounding-box, and $v$ is the ground truth of those offsets. $\lambda$ is a hyper-parameter used to control the balance classification loss and regression loss.

Most of two-phase WSOD architectures utilize Fast RCNN as the second-phase detector. As a baseline, we also adopt Fast RCNN as one of the second-phase detectors in our work.

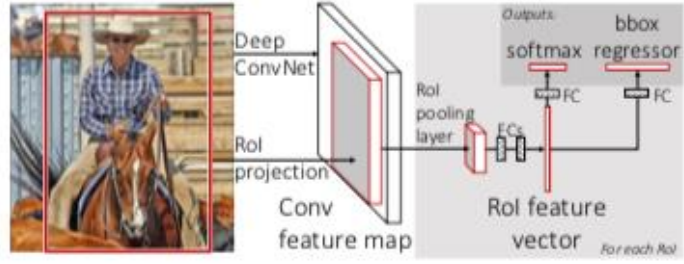

Figure 3. The architecture of Fast RCNN. The inputs are images and their corresponding region proposals. Through CNN, proposal features are created which are then mapped to fixed-size feature vectors via ROI pooling and FC layers. The outputs comprise softmax probabilities and per-class bounding-box regression offsets.

## 2.2 Faster RCNN

Faster RCNN [11] is the successor of Fast RCNN and provides great success in object detection tasks. The purpose of Faster RCNN is to improve the precision and speed of Fast RCNN. Fast RCNN requires precomputed region proposals to train the network. However, a large amount of time is consumed to create these region proposals. Hence, to deal with this problem, designers of Faster RCNN intend to generate region proposals in the network to establish a unified architecture. This is implemented by integrating Region Proposal Network (RPN) into the Fast RCNN. The function of RPN is to create a large number of object proposals and their corresponding objectness scores. Faster RCNN further improves the speed of Fast RCNN because there are many shared convolutional (conv) layers between RPN and Fast RCNN. Therefore, it is efficient to generate region proposals and train the detector in this unified architecture compared with Fast RCNN. Specifically, following the last shared conv layer, the RPN is utilized to generate region proposals. Region proposals with high confidence of being an object are then sent to Fast RCNN to perform classification and bounding-boxes regression over all classes. To jointly train these two networks, they propose a four-step algorithm. (1) Fine-tune a RPN intialized with ImageNet-pre-trained model for region proposal task. (2) Initialized with an ImageNet-pre-trained model, a Fast RCNN is trained according to the region proposals generated by (1). (3) Fine-tune the layers unique to RPN initialized with the Fast RCNN trained in (2), while the shared conv layers are fixed. From this step, two networks share conv layers. (4) Fine-tune fully connected layers of Fast RCNN.

Faster RCNN implements RPN by a small network over the last shared conv layer. By sliding this network over the conv feature map created by the last shared conv layer, region proposals are generated. To be specific, there are several reference boxes (anchors) at each sliding window position. Region proposals are created via mapping the anchors to the original image according to the sliding window location. The structure of RPN is shown in Fig. 4 [11]. The loss function of RPN is demonstrated in Equation (6) [11]. Due to the high-quality generated by RPN, Faster RCNN also outperforms Fast RCNN in detection accuracy.

$$L(\{p_i\}, \{t_i\}) = \frac{1}{N_{cls}} \sum_i L_{cls}(p_i, p_i^*) + \lambda \frac{1}{N_{reg}} \sum_i p_i^* L_{reg}(t_i, t_i^*) \qquad (6)$$

in which $L_{cls}$ is the log loss over two classes and $L_{reg}$ is shown in Equation (7).

$$L_{reg} = smoot_{L1}(t_i - t_i^*) \qquad (7)$$

Where $i$ denotes the index of an anchor in a mini-batch, $p_i$ is the predicted probability of anchor $i$ being an object, and $p_i^*$ is the ground truth label belonging to {0, 1}(1 for positive and 0 for negative). $t_i$ refers to the four coordinates of predicted bounding boxes and $t_i^*$ is the corresponding ground truth. $N_{cls}, N_{reg}$ are normalization factors and the function of $\lambda$ is the same as Fast RCNN mentioned before.

Although great success can be seen in object detection in fully supervised Faster RCNN, few studies have explored the performance of the Faster RCNN as the second-phase detector in WSOD. This project will investigate the performance of this architecture.

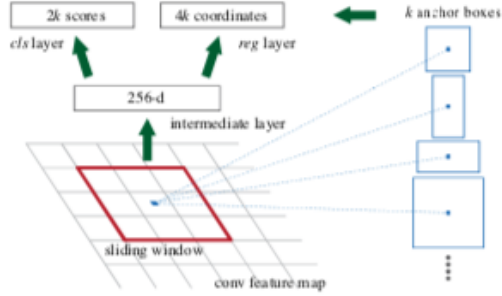

Figure 4. The structure of RPN. For each sliding window location, there are $k$ predicted region proposals according to the $k$ anchor boxes. The features of these proposals are then classified to two classes {object, non-object}. Hence, there are $2k$ scores. Each region proposal requires four coordinates, thus there are $4k$ coordinates.

## 2.3 Multiple Instance Learning

Proposed by Dietterich et al. [9], MIL is designed to solve weakly supervised problems and was first applied to drug activity detection [15]. In MIL, the main and basic component is a bag containing a collection of instances. A positive bag at least contains one positive instance, otherwise it is a negative bag. Only given the bag-level label, the aim is to train a multiple instance classifier and apply this classifier to a new dataset. It is very natural to link MIL with weakly supervised computer vision tasks because of their high similarity. In WSOD, only image-level supervision given, the instance is the region proposal and we intend to obtain the category and bounding boxes of each potential object. To be specific, training images are divided into positive and negative bags according to whether they contain a specific object class. These bags are then used to train a detector to learn discriminative representation of different object instances. However, it is more complex to apply MIL to WSOD as object detection tasks require outputs comprising of both category and bounding boxes but bounding boxes are not available during training, which reduces the performance of MIL. Fortunately, an abundance of research has proposed methods [8, 15, 17, 18] to alleviate this problem and exploits MIL to solve the WSOD. Details of these methods are introduced in Section 2.4 and Section 2.5.

## 2.4 Weakly Supervised Object Detection

Object detection aims to detect instances of semantic objects of some certain classes, such as pedestrians, cars and buildings. State-of-the-art methods are normally trained with instance-level annotations. During the training of these methods, the class labels and their coordinates in images are given. Thus, models are always supervised by the right direction and information. Nonetheless, it is of great difficulty from an economic perspective to build those datasets containing bounding box annotations since this requires people to manually annotate the objects and their positions in each image. Therefore, many researchers explore the object detection with only image-level labels and gain significant improvement. One of our objectives is to integrate some of these models with representative and powerful networks to test whether this architecture can further boost the performance.

The main issue of WSOD is that it is much harder to generate the bounding box of objects as no ground truth location is available. In WSOD, the model may be misled and this may be amplified if the supervision given is always wrong. To solve the WSOD problem, current methods normally adopt the MIL strategy and take advantage of CNN, and form an end-to-end unified learning approach.

Weakly Supervised Deep Detection Network (WSDDN), proposed by Bilen et al. [1] is a two-stream architecture. One stream for classification and one stream for detection. The positive sample is determined in terms of the score of these streams. Based on WSDDN, Kantorov et al. [8] adds context information and proposes a contextual modelling. One problem that these methods have is that the algorithm is likely to be trapped into local minimum. To alleviate this problem, Tang et al. [16] presents an online instance classifier refinement (OICR) algorithm which is also based on WSDDN. Also proposed by Tang et.al. [15], Proposal Cluster Learning (PCL) clusters the region proposals of an image into different clusters before sending them as the input to the backbone network. In PCL, instead of assigning each region proposal a class label, they treat each cluster as a bag and design a loss function based on this, making the algorithm pay attention to the whole part of object, and thus, reduce the ambiguity. An attention mechanism is integrated into [18] to further improve the model by automatically assigning the weight for different features.

Current state-of-the-art algorithms of WSOD follow a two-phase learning strategy. In this strategy, a weakly supervised network (e.g. methods mentioned above) is first trained to obtain the predicted bounding boxes on training dataset. These predicted bounding boxes are then regarded as the PGT to fine-tune a second-phase detctor. [15, 17, 19, 20, 21] exploit Fast RCNN as the second-phase network and outperform the one-phase models. In this project, we will also adopt Fast RCNN as our second-phase detector to verify the effectiveness of our methods because Fast RCNN is the baseline network in object detection. As the successor of Fast RCNN, Faster RCNN outperforms Fast RCNN both in precision and speed in fully supervised scenario, and thus we exploit it as the second-phase detector and verify whether this can further improve the accuracy in WSOD. The principle of Fast RCNN and Faster RCNN are detailed in Section 2.1 and 2.2. PCL is utilized as the first-phase detector since it is the state-of-the-art one-phase WSOD model. We will introduce the details of PCL in Section 2.5.

## 2.5 Proposal Cluster Learning

Our project adopts PCL as the first-phase detector. Therefore, it is essential to clearly comprehend PCL before introducing our methods. Proposed by Tang et al. [15], Proposal Cluster Learning is the state-of-the-art single network model in WSOD. Like the projects mentioned in Section 2.3, PCL also integrates MIL constraints to transfer the instance classification (object detection) problem to the bag classification (image classification) problem. Nonetheless, they observe that there is a great difference between object detection and image classification tasks. Many region proposals are a part of the whole object. For image classification, even only a small patch of objects may be enough to provide the right result as these parts normally contain distinguishable information. For instance, the head of a person allows the classifier to predict the right decision. However, this characteristic does not have the same effect on object detection because object detection tasks not only require algorithms to classify the category, but also localize objects with bounding boxes properly. In terms of the evaluation standard Intersection-over-Union (IoU), parts of objects classified and detected correctly are not sufficient for the task requirement. To alleviate this problem, PCL clusters region proposals during training to focus more attention on the larger part and develops a multi-stream output network to refine the results. The details will be introduced in Section 2.5.1 and 2.5.2. We describe the proposal cluster and the approach of PCL to produce it in Section 2.5.1. The architecture and training process are interpreted in Section 2.5.2.

### 2.5.1 Proposal Cluster

In Region-based CNN object detection, the algorithm first generates region proposals and expects them to have high IoU with the ground truth to guarantee the high proposal recall. Classification is then performed on these region proposals and regression may be required to fine-tune the position of proposals. Current deep learning object detection algorithms normally train a detector based on these proposals. Although they have achieved great success, there are some potential problems that limit the performance. As stated in Section 2.5, most of these proposals only cover a part of an object. In addition, the detector may be confused since different parts of an object may lead to the same classification result.

To alleviate the problems mentioned above, PCL proposes to cluster the region proposals. A proposal cluster is a group of proposals having high spatial similarity with other proposals in the same group. For example, there are multiple objects from multiple categories in an image and the corresponding proposals of each instance are a proposal cluster. Backgrounds of images are considered as a special proposal cluster since no cluster centre exists in this cluster. Within these proposal clusters, detectors can focus on larger parts during training by assigning proposals in each cluster the same label as the object class.

The main problem then is how to generate these clusters. In a fully supervised scenario, ground truth bounding boxes are regarded as cluster centres. Based on these cluster centres, images can be simply grouped into different clusters via comparing their IoU with cluster centres. However, instance-level information is not available in WSOD, and thus exploiting proposal cluster strategy in WSOD is difficult. To deal with this problem, PCL proposes a graph-based approach to select cluster centres. Specifically, top score proposals of each class are chosen as corresponding cluster centre candidates according to the proposal scores calculated by the algorithm. They then build a graph on these candidates. In this graph, each proposal in the candidate set is a vertex. For the generation of edges, if the IoU between any two proposals exceeds a threshold (e.g. 0.4), there is a connection (edge) between these two vertexes. After obtaining the graph, the algorithm selects cluster centres through the following steps. Firstly, the vertex with the most connections (edges) is selected as one cluster centre. The algorithm then removes vertexes having connections to the cluster centres generated in step one and their corresponding connections to other vertexes. Repeating the above two steps until the vertex set is empty, we can then collect the cluster centres. Following this method, the number of cluster centres may differ in training stages as proposal scores in different images and training step are different, which deals with the case that multiple instances from the same class occur in one image. After obtaining the proposal clusters, the problem is then looking at how to train a detector according to proposal clusters. This is introduced in Section 2.5.2.

**2.5.2 Architecture and Training**

In order to take advantage of the proposal clusters mechanism, PCL presents a bag-based loss function and a multiple output network architecture shown in Fig. 5 [15]. In the left part of this figure, a traditional CNN network is exploited to create proposal features. These proposal features are then fed into the first stream, a basic MIL network which transforms object detection problems into image classification problems and is aggregated to image scores to train a basic instance classifier. After that, this classifier will be further fine-tuned by the subsequent streams. Built in the proposal scores obtained in the first stream, proposal clusters are generated and the instance classifier will then be fine-tuned according to these proposal clusters. To be more general, assuming that we have $K + 1$ streams, the proposal scores in $K$ stream and their corresponding image labels are considered as the supervision in the $K + 1$ stream to train the instance classifier. These streams share the proposal features, and thus, reduce the computation complexity. To implement their idea that training a network according to proposal clusters, they propose a bag-based loss function as shown in Eq. (8) [15]. A proposal cluster is regarded as a new bag. Instead of proposal-level labels, they utilize bag-level labels in the loss function since this can alleviate the problem where proposals containing different parts of objects with the same class label may confuse the classifier.

$$L^k(F, W^k, H^k) = -\frac{1}{R} \left( \frac{\sum_{n=1}^{N^k} S_n^k M_n^k \log \sum_{r \, s.t \, b_r \in B_n^k} \varphi_{y_n^k r}^k}{M_n^k} + \sum_{r \in C_{N^k+1}^k} \lambda_r^k \log(\varphi_{(C+1)r}^k) \right) \qquad (8)$$

$k$ denotes that we are calculating the loss for $K$ stream or the parameters are utilized for $K$ stream. For example, $L^k$ means that we intend to calculate the loss for $K$ stream. $R, C, N^k, S_n^k$ and $M_n^k$ are the number of proposals, the number of classes, the number of object clusters in $K$ stream, the cluster confidence score of the $n$-th proposal cluster and the number of proposals in the $n$-th proposal cluster respectively. $C + 1, N^k + 1$ are for background proposal cluster. We denote the predicted score of a proposal $r$ for class $c$ in stream $k$ as $\varphi_{cr}^k$. $S_n^k$ is set to the highest score in the proposal cluster. $B_n^k, y_n^k$ is the bag for $n$-th proposal cluster and the bag label. $\lambda_r^k$ is the proposal scores of proposal $r$.

In testing time, final detection scores are calculated by averaging the proposal scores of all refined instance classifiers as shown in the blue arrows in Fig. 5 [15].

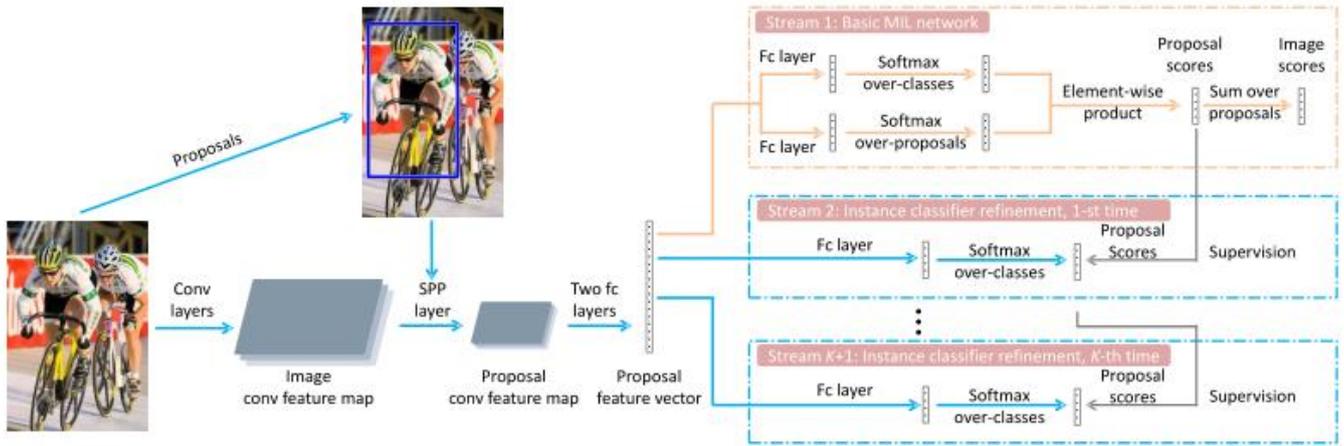

Figure 5. The architecture of PCL. All arrows are used during the forward-process of training while only solid arrows perform back-propagation. In testing time, only blue arrows are utlized. During training, a CNN network is utilized to extract image features. These features combined with pre-computed proposals boxes are then sent to the SPP layer to generate proposal features. These proposal features are then sent to two fully connected layers to compute the classification and detection scores respectively in the first stream. The instance classifier is refined by the following streams based on the proposal clusters. All streams share the same proposal features.

# 3 METHODOLOGY

This section describes our proposed methods to advance the precision of WSOD. We improve the performance of original PCL from three directions corresponding to objective one to three. (1) Select Fast RCNN and Faster RCNN as the second-phase detector. (2) Adopt more bounding boxes to generate the PGT. (3) Present multiple strategies to refine PGT during the training of the second-phase detectors. Implementation Details are introduced in Section 4.

## 3.1 Two-phase WSOD

PCL has proven that a Fast RCNN can boost the performance of their ensembling method. However, no experiments are set to their single model. In addition, in the newest implementation of PCL, they propose two tricks and gain significant improvements with their precision. Considering these two conditions, this project explores whether a Fast RCNN can further improve the performance in their newest single network implementation. Moreover, we utilize a more powerful detector, Faster RCNN, and expect this can increase precision.

To implement our idea, a PCL network is first trained with only image-level annotations. This means that bounding boxes of each instance and the number of instances for each class in the image are not available during the training of PCL. Hence, algorithms only have the knowledge about what categories exist in the image. Built in the trained PCL model, the algorithm then performs inference on the training dataset to obtain the predicted bounding boxes and their categories. Next, we establish instance-level annotations to train the second-phase model according to the results obtained in the previous step. During the training on the second-phase network, algorithms know the number of instances in each class and their coordinates information in images, but this information is created by the first-phase network. Hence, it is the PGT. After training, the performance of this two-phase WSOD architecture will be verified through testing the second-phase detector on the test dataset. One significant procedure in this architecture is how to establish the PGT for the second-phase network from the the first-phase model. This is detailed in Section 3.2. The datasets and evaluation metrics we used to verify the performance are illustrated in Section 5.1 and 5.2 respectively.

## 3.2 PGT Generation

PGT is the core to connect the first-phase and second-phase models. As stated before, most previous works select top one bounding boxes as the PGT to train a second-phase model such as Fast RCNN. It is very natural to select the most confidence bounding box since detectors tend to be trained with correct supervision according to these high score boxes. Nonetheless, this could lead to some problems affecting the performance of the detector. (1) We empirically observe that objects from the same class may appear more than once in a large number of pictures and datasets. In this case, only given the top one bounding boxes, the detector may be unable to learn adequate features to distinguish this class from others. This problem could be deteriorated when the dataset is small or there are many subclasses in a class such as Border Collie and Akita in the dog category. (2) In WSOD, a relatively high probability can be seen that top bounding boxes are misclassified or contain only parts of an object. These cases may also cause an insufficient learning problem.

To mitigate the problems mentioned above, we propose to consider top $k$ bounding boxes as the PGT to guarantee a high proposal recall. Nonetheless, a large $k$ value may also introduce many negative samples, and thus mislead the learning while a small $k$ value may lose significant learning information. Hence, it is essential to maintain the balance between high proposal recall and high precision. In our work, we elaborately conduct many experiments to test the performance on different settings of $k$ on the PASCAL VOC2007 datasets. $k$ is set to $\{1, 2, 3, 5, 50\}$. One limitation of this method is that the values of $k$ demonstrating best performance may differ in datasets since the object distributions of datasets are different. Although larger $k$ values may not gain acceptable precision, we also conduct some experiments with very large $k$ values (e.g. 50) since the results may deliver implications to the future works (e.g. regenerate more accurate PGT based on the increment of proposal scores).

We collect top $k$ bounding boxes in three steps. 1) We train a PCL model on a training dataset, and then perform inference on the same dataset to obtain the predicted bounding boxes and their scores. 2) Recall that algorithms do not know the number of instances in each category in images and their bounding boxes but they do have knowledge about which class occurs in the image for WSOD. We record this information via processing the annotation file. 3) Based on information obtained from step one and two, top $k$ bounding boxes are created by ordering the bounding boxes based on their scores in corresponding categories and selecting the best $k$ boxes. We then save the PGT to establish an instance-level annotation file which will be processed to train the second-phase model. We summarize this procedure via pseudocode in Algorithm 1. $B$ is the predicted bounding boxes generated by PCL and $S$ are their socres. $R$ is the number of predicted bounding boxes for $i^{th}$ image generated by PCL. $N$ is the number of images in the training set.

---

**Algorithm 1** Creating PGT

**Input**: Training images: $I$
 Trained PCL model: PCL
 Classes occurs in each image : $C_i$.
 The number of top bounding boxes: $k$
**Output**: $PGT$
1. Initialize $PGT = \emptyset$
2. $B, S$=PCL($I$)
3. **For** $i$ = 1 to $N$, **do**
4.   **For** $j$ in $C_i$, **do**
5.     $B_i$ =Sort($B_i, S_{ij}$)
6.     $PGT_i$=Top($B_i, k$)
7.     $PGT$.append($PGT_i$)
8. Save $PGT$ and update annotation documents

---

## 3.3 PGT Refinement

In this section, we develop a method to refine the PGT during the training of the second-phase network. Given the PGT, it has been proven that some models can improve the results from a pure WSOD model due to their powerful feature extraction ability and architectures. However, for those misclassified samples and samples only containing non-distinguishable parts, a second-phase detector seems powerless. Considering this, we propose a method to refine the PGT during the training. This strategy can improve the PGT or increase the number of positive training samples, and thus may boost the performance. The reason is based on the fact that weights of positive samples will be enhanced when training detectors in other images containing the same features while weights of negative samples may be learned slowly if the positive samples outnumber the negative ones. To implement our idea, we perform inference on the training dataset during training and recreate instance-level annotations according to the bounding boxes scores. Unlike the PGT generation in Section 3.2, this update is performed during training. This means that the weights and gradients of the network are not changed after performing inference on training dataset. There are two reasons for choosing to adopt this strategy. (1) Compared with finishing training and then refining the PGT in a new model, this strategy reduces the calculation complexity. (2) During the training, the detector may not be misled from those negative training samples much, or unsaturated. On the other hand, if we utilize a same model to refine the PGT given by the trained second-phase detector, this means that we perform a three-phase WSOD task. The performance may not be improved due to the high similarity between the second-phase and third-phase network.

We propose seven strategies to implement our idea. These strategies are divided into two groups as shown below. Strategies in group one determine the time to refine the PGT, while those in group two define how to update the PGT. A complete refinement strategy comprises one strategy from group one and one strategy from group two. We verify their effectiveness on the PASCOL VOC2007 dataset with Faster RCNN as the second-phase model. The components and evaluation metrics of the PASCOL VOC2007 dataset are demonstrated in Section 5.

Group 1:
- Refine the PGT after each epoch.
- Refine the PGT every three epochs。
- Refine the PGT after each epoch in the last three epochs.
- Refine PGT only once at the 2/3 of the maximum epochs.

Group 2:
- Update all $k$ PGT combing with strategy in group one.
- Only update half of the $k$ best PGT combining with strategy in group one.
- Only update half of the $k$ worst PGT combining with strategy in group one.

To be specific, we do this in the following steps. As a necessary procedure, a PCL model is trained to generate the annotations (PGT) for the second-phase detector via the method described in Section 3.2. We then train the second-phase model based on the PGT. At specific epochs during the training (e.g. epoch mentioned in group one), we suspend training, save the model, and change the model from training mode to evaluation mode. After suspending the training, we perform inference on training images and then update the annotation files according to the strategies mentioned in group two. One thing that should be emphasized is that we do not stop the training, and thus, the gradient and the weight of the network are kept. The next step is to continue training accroding to the refined annotations.

# 4 IMPLEMENTATION DETAILS

This section clearly introduces the backbone network and programming environments used in this project. In addition, our work is established based on some open-source projects, and thus, they are also briefly described in this section.

## 4.1 Backbone network

We select ResNet[14], combined with Feature Pyramid Network (FPN) [22], as the backbone networks for Fast RCNN and Faster RCNN due to their significant achievement in Computer Vision tasks. ResNet is the state-of-the-art backbone model and maintains competitive performance in many tasks such as image classification and object detection. It enables very deep CNN training via adding residual block to deal with the gradient vanishing problem. In object detection, multi-scale training is significant because it is common that the same objects may occur in multiple scales. FPN introduces multi-scale factor and enhances multi-scale training ability of algorithms.

## 4.2 Programming Language

This project utilizes Python as the programming language. There are four reasons on why we have chosen Python. (1) Python has a strong programming ability and supports many high-level interfaces while the semantics are not complex. (2) Python has a large number of excellent libraries that are widely used in data analysis such as numpy, pandas, and matplotlib. By taking advantage of these libraries, we are able to implement our idea efficiently. (3) Our work is based on previous established works which are implemented via python. (4) Excellent community support.

## 4.3 Anaconda

Anaconda, an open-source package manager of Python for scientific computing, aims to simplify package management and deployment. Through Anaconda, users can create an environment and install all required packages via command line in this environment. It is very convenient since users do not need to download packages and install them manually. Moreover, each environment is independent and has no influence on other environments. This project must deploys environments to train PCL, Fast RCNN and Faster RCNN. However, environments of these three models are different (e.g., different python version), and thus we cannot install them in a base environment. Therefore, these reasons encourage us to take advantage of Anaconda as the package manager.

## 4.4 PyTorch

PyTorch is a well-known open-source deep learning framework. It provides a lot of friendly deep learning interfaces and functions such as network layers and data preprocessing tools. Besides, it also supports multi-gpu training and cloud platforms. PyTorch has been widely used in deep learning research and has an active community. Through PyTorch, we can establish, train, and test a deep learning network efficiently and effectively. Hence, PyTorch is selected as the fundamental deep learning framework in this project.

## 4.5 Detectron2

Developed and maintained by Facebook AI Research, Detectron2 is an open-source project which implements some state-of-the-art object detection models. This open-source project is powered by the PyTorch deep learning framework, which is another reason for us to select PyTorch. It trains networks much faster and can be used as a library to support different projects on top of it. Many works establish their projects on Detectron2 and share their experiences and questions on forums. The developers of Detectron2 also provide complete and detailed documentations. These two help researchers implement their projects efficiently.

As stated before, we utilize Fast RCNN and Faster RCNN as the second-phase detector in our project. Detectron2 has developed Fast RCNN and Faster RCNN and achieved the accuracy reported in papers. Hence, implementing our ideas on Detectron2 could be simple and fast. Considering these advantages, we establish our project on Detectron2 via modifying a section of source code and adding our code to implement our methods. Detectron2 requires many libraries to support its operation and performance. A GPU is used to accelerate the computation. The basic software requirements and their version in our work are shown in Table 1. One thing that should be emphasized is that version differences may lead to the failure of Detectron2 establishment.

Table 1. Basic software requirements for detectron2.

| Software requirements | Version | Software requirements | Version |
|---|---|---|---|
| python | 3.7.7 | pytorch | 1.4.0 |
| cython | 0.29.0 | matplotlib | 3.2.2 |
| numpy | 1.18.1 | opencv | 3.4.2 |
| pillow | 7.1.2 | pyyaml | 5.3.1 |
| pycocotools | 2.0.1 | pip | 20.1.1 |
| tensorboard | 2.2.2 | torchvision | 0.5.0 |

## 4.6 ElementTree

ElementTree is a package developed by the Python developer team. This package implements a simple and efficient interface for parsing and creating Extensible Markup Language (XML) data. The annotation file of PASCAL VOC2007 datasets is formed via XML. In our project, the algorithm reads and modifies annotation files frequently to generate PGT in different $k$ settings. Hence, an effective XML processing tool is important. ElmentTree parses a XML document as a tree structure, and it is convenient and simple to iterate each tag in a tree structure. Due to the friendly and efficient interface of ElementTree, we utilize it to process XML documents.

## 4.7 PCL Environments

As we mentioned in the previous section, the environment to train PCL is different from Fast/Faster RCNN (Detectron2). Like Detectron2, we list the software requirements and their version in Table 2. PCL and Detectron2 share some packages, but their versions may be different. Installing these packages in one environment may lead to potential conflicts.

Table 2. Basic software requirements for PCL.

| Software requirements | Version | Software requirements | Version |
|---|---|---|---|
| python | 3.7.7 | pytorch | 1.4.0 |
| cython | 0.29.0 | matplotlib | 3.2.2 |
| numpy | 1.18.1 | opencv | 3.4.2 |
| pillow | 7.1.2 | pyyaml | 5.3.1 |
| pycocotools | 2.0.1 | pip | 20.1.1 |
| tensorboard | 2.2.2 | torchvision | 0.5.0 |

## 4.8 Hardware Configuration

This section introduces the hardware configuration of our computer that is performing the experiments. Due to the heavy computation load and elaborate experiments with limited time, a GPU is required to accelerate computation. This configuration is a reference for those who intend to reproduce our experiments. Other configurations may also be feasible. The hardware configurations for Fast/Faster RCNN and PCL are demonstrated in Table. 3.

Table 3. Hardware configurations for Fast/Faster RCNN and PCL.

| Hardware | Fast/Faster RCNN | PCL |
|---|---|---|
| Operating System | Ubuntu 18.04 | Ubuntu 16.04 |
| CPU | Intel Core(TM) i9-10900X | Intel Xeon(R) CPU E5-2670 |
| GPU | RTX 2080 Ti 11G | GTX 1080 Ti 11G |
| Memory | 128G | 64G |

# 5 EXPERIMENTS AND RESULT ANALYSIS

In this section, we first introduce the dataset and evaluation metric. Augmentation methods utilized in our experiments are then described. Next, we conduct elaborate experiments to explore the performance of our methods and different settings. Finally, we compare our best performed method with other methods and demonstrate some limitations of our work.

## 5.1 PASCAL VOC2007 Dataset

The PASCAL VOC2007 dataset is a benchmark to measure the ability of a model in image classification and recognition. A large amount of works test their models and report the results on this dataset. Hence, to provide comparisons, this project also utilizes the PASCAL VOC2007 dataset to train models and verify the performance of our methods. This dataset consists of 5011 training images and 4952 testing images. These images come from 20 classes and are saved in the JPEG format. Their categories and amounts on training and testing set are shown in Table 4 and Table 5 respectively. The annotation files are formed by Extensible Markup Language (XML). For object detection tasks, annotations contain categories and their bounding boxes of all instances in each image. Since this project focuses on WSOD, the bounding box tag is neglected when processing the annotation file.

Table 4. The amount of instance from each category in training set.

| Category | Account | Category | Account |
|---|---|---|---|
| aeroplane | 238 | diningtable | 200 |
| bicycle | 243 | dog | 421 |
| bird | 330 | horse | 287 |
| boat | 181 | motorbike | 245 |
| bottle | 244 | person | 2008 |
| bus | 186 | pottedplant | 245 |
| car | 713 | sheep | 96 |
| cat | 337 | sofa | 229 |
| chair | 445 | train | 261 |
| cow | 141 | tvmonitor | 256 |

Table 5. The amount of instance from each category in testing set.

| Category | Account | Category | Account |
|---|---|---|---|
| aeroplane | 204 | diningtable | 190 |
| bicycle | 239 | dog | 418 |
| bird | 282 | horse | 274 |
| boat | 172 | motorbike | 222 |
| bottle | 212 | person | 2007 |
| bus | 174 | pottedplant | 224 |
| car | 721 | sheep | 97 |
| cat | 322 | sofa | 223 |
| chair | 417 | train | 259 |
| cow | 127 | tvmonitor | 229 |

## 5.2 Evaluation Metrics

The PASCAL VOC2007 Dataset adopts Mean Average Precision (mAP) as its evaluation metrics. Hence, it is of great significance to comprehend the mAP before delivering the experiments and results analysis.

We first introduce the definition of precision, recall and how to calculate them. We then interpret Average Precision(AP) for object detection tasks. mAP is the mean value of AP over all classes.

**5.2.1 Precision and Recall**

Precision refers to the proportion of predicted results which are truly correct while recall describes the proportion of instances that are correctly detected. Predicted results of a model are divided into true positive (TP), false positive (FP), true negative (TN) and false negative (FN). TP, FP, and FN are three terms used to calculate mAP for object detection. An instance is correctly detected if the IoU between its predicted bounding boxes and the ground truth is larger than a threshold (e.g. 0.5). For object detection tasks, positive samples are predicted bounding boxes with confidence larger than the threshold. True positives are positive samples that correctly detecting objects, otherwise they are false positives. As a special case, duplicate predicted bounding boxes are also considered as FP. False negatives refer to instances that are ignored (no predicted bounding boxes) by the algorithm. We do not introduce TN as it is not used in the calculation of AP for object detection tasks. After comprehending the definition of these terms, precision and recall are calculated via Eq (9) and (10). However, a single indicator (precision or recall) normally cannot completely evaluate the performance of an algorithm. For example, all predicted results of a model are correct, a full mark precision, but this model ignores many instances, a very low recall. Hence, this is not a successful model. To cope with this problem, precision-recall curve is a commonly and widely used indicator to evaluate models.

$$precision = \frac{TP}{TP+FP} \qquad (9)$$
$$recall = \frac{TP}{TP+FN} \qquad (10)$$

To clearly illustrate these three terms, an example is provided and shown in Fig. 6. Red bounding boxes are ground truth. Blue bounding boxes (positives) are predicted bounding boxes with confidence larger than the threshold given by the algorithm. Box one is a TP because its IoU is larger than 0.5, while box two is a FP as it does not fit the ground truth properly. In addition, the cat in the right half of the image is not detected by the algorithm, and thus it is a FN. After obtaining the value of the TP, FP and FN, the precision and recall are then calculated via Eq (9) and (10). Table 6 presents the results of these terms.

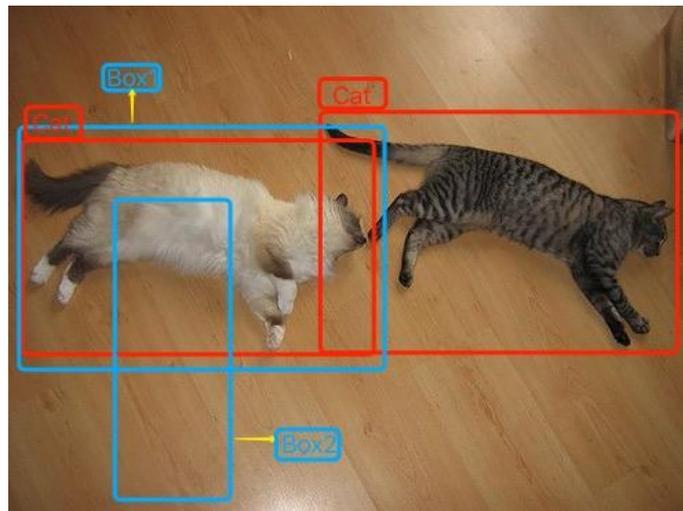

Figure 6. An image with predicted bounding boxes and ground truth. Red bounding boxes are ground truth. Blue bounding boxes are predicted bounding boxes with confidence larger than a threshold given by the algorithm.

Table 6. Results of TP, FP, FN, and precision as an example.

| TP | FP | FN | Precision | Recall |
|---|---|---|---|---|
| 1 | 1 | 1 | 1/2 | 1/2 |

### 5.2.2 Average Precision

Average precision summarizes the shape of precision-recall curve. Although precision-recall curve is widely used as indicators to evaluate an algorithm, they cannot simply be applied to object detection models. The main reason is that results returned by an algorithm are a sorted list according to their confidence. As we mentioned above, positives are predicted bounding boxes with confidence larger than a threshold. Different algorithms may choose different thresholds, and thus lead to different precision-recall curves. To deal with this problem, PASCAL VOC 2007 requires the outputs to be sorted first according to their confidence, and then calculates AP on a precision-recall curve. AP for the VOC 2007 dataset is the mean precision at eleven equally spaced recall levels [0, 0.1, …, 1]. It is calculated by Eq (11) and (12). An interpolation strategy is adopted in precision calculation to reduce the influence of the "wiggles".

$$AP = \frac{1}{11}\sum_{r\in\{0,0.1,…,1\}} P_{interp}(r) \quad (11)$$
$$P_{interp}(r) = \max_{\tilde{r}:\tilde{r}>r} p(\tilde{r}) \quad (12)$$

### 5.3 Data Augmentation

Data augmentation is a technique to increase the diversity of data without collecting new data. A large neural network normally requires a large dataset to ensure complete training. Strategies widely utilized in computer vision tasks are cropping, padding and horizontal flipping. In this work, for the training phase, we first perform normalization on each image to eliminate the dependency between learning rate $lr$ and the input images. Horizontal flipping is then utilized on images to enrich datasets both in training and testing time. In addition, like [15], we use five image scales {480, 576, 688, 864, 1200} to enable multi-scale training via resizing the shortest side to the closest scales. Only Faster RCNN supports multi-scale testing since Fast RCNN requires pre-computed region proposals to perform training and inference. If we enable multi-scale testing on Fast RCNN, the algorithm should generate four more region proposals for each original region proposal and then perform inference on these region proposals. However, there are approximately 2K region proposals per image in the original region proposal set. Hence, it consumes a large amount of time to perform inference on images on Fast RCNN with multi-scale testing time augmentation. Considering our computer resources and time limitation, it is unrealistic for us to enable the multi-scale strategy on Fast RCNN.

### 5.4 Experimental setup

Elaborate experiments are conducted to verify the performance of our three methods. We present the details about the experimental groups in this section. Fast/Faster RCNN exploit the same backbone network FPN while the depth of another backbone network ResNet may be different. For simplification, we do not mention the backbone network FPN when we list the backbone networks of Fast/Faster RCNN in the following sections. (1) For two-phase WSOD, the first-phase model is PCL and the second-phase detectors are Fast RCNN and Faster RCNN. The reasons for selecting these models are introduced in Section 3. Hence, two groups of experiments are set to verify the performance of these two second-phase detectors. In this part, we select top one bounding boxes generated by PCL as PGT to train the second-phase detectors. As mentioned in Section 4.1, ResNet is adopted as our backbone network for Fast/Faster RCNN. There are three different depths of commonly used ResNet {ResNet34, ResNet50, ResNet101}. To investigate whether a deeper architecture can further improve the performance while ensuring the acceptable precision, we select ResNet50 and ResNet101 pretrained on ImageNet. (2) For PGT generation, we conduct experiments on six values of $k$ {1,2,3,4,5,50}. We only utilize Faster RCNN as the second-phase detector because it obtained great results in (1). Therefore, there are six groups of experiments corresponding to six $k$ values. (3) For PGT refinement, we only use strategies on Faster RCNN, because the aim of this section is to verify the effectiveness of refinement strategies introduced in Section 3.3 and Faster RCNN outperforms Fast RCNN. Moreover, some strategies only update half of the $k$ bounding boxes, thus we select $k$ {2} in this section to ensure that the number of bounding boxes that are updated is an integer. Recall that there are two groups of strategies, four strategies for group one and three strategies for group two, and a complete refinement strategy consists of one strategy from group one and one strategy from group two. Hence, **4 ∗ 3** experiments are set. Results and analysis are demonstrated in next section.

We train all models with $12K$ iterations and the image batch is set to 4. The initial learning rate $lr$ is 0.002 and is decayed in 8000 and 10500 iterations.

## 5.5 The performance of two-phase WSOD

Here, we compare our method with the original PCL. Recall that we train a Fast RCNN and a Faster RCNN to improve the results generated by PCL. The annotation files for second-phase detectors are obtained by considering the top one bounding box as PGT. As stated in Section 3.1, the PyTorch implementation of PCL embeds two tricks and gains significant improvement while the original implementation is based on Caffe. Our work utilizes PyTorch version PCL due to its powerful performance. We report our results and PCL (two types) on the VOC 2007 test set in Table 7. It is emphasized again that no multi-scale strategy is applied to the testing of Fast RCNN, and this strategy may increase the mAP by approximately 2.5% to 3.5% according to the experiments in Faster RCNN and statistics reported in PCL. From Table 7, we can see that there is no obvious improvement for selecting Fast RCNN as a second-phase detector when we add the maximum possible increment obtained from multi-scale strategy compared with the PyTorch implementation of PCL. One possible reason is the high similarity between PCL and Fast RCNN. These two models train the networks based on the same precomputed region proposals and similar CNN architecture, which may lead to similar proposal features. Learning on these proposal features with annotations generated by PCL is prone to produce similar results. Moreover, some of these PGT are negative samples, thus the majority of region proposals sampled in the mini-batch are also negative samples in Fast RCNN. Hence, compared with the weight of positive samples, the weight of these negative PGT may be considerably increased when training on these negative region proposals, which limits the performance.

Nonetheless, a different pattern can be seen in Faster RCNN. As shown in Table 7, Faster RCNN gains approximately 4% mAP improvement from 49.17% to 53.21%. This confirms the effectiveness of our PCL-Faster RCNN architecture. Furthermore, Faster RCNN almost obtains the best AP in all classes except for bus, motorbike, person, and sofa. This improvement is mainly obtained due to the high-quality proposals created by the Region Proposal Network in training and testing images. These high-quality region proposals normally have high IoU with PGT and are likely to contain at least parts of ground truth. Especially in testing time, no ground truth annotations are available. Therefore, although the Faster RCNN also suffers from the same misleading problem as Fast RCNN, the influence of this problem is alleviated. In addition, given high-quality region proposals, powerful image recognition abilities of Faster RCNN is another reason for the better performance.

As stated in Section 3.1, we intend to investigate the performance of a deeper architecture compared with the shallow one, thus conducting experiments on ResNet101 and ResNet50. From the third to the last row of Table 7, the mAP increases slightly from 44.739% to 45.063% and 53.163% to 53.212% for Fast RCNN and Faster RCNN respectively. It is obvious that increasing the depth of the network does not significantly improve the performance for both Fast RCNN and Faster RCNN. This indicates that ResNet50 is sufficient to extract distinguishable features on VOC 2007 under this setting.

Table 7. Results (AP in %) for different methods on VOC 2007 test set. PCL results reported in this paper utilize single VGG16 model. Res50 denotes ResNet50 and Res101 denotes ResNet101.

| Method | aero | bike | bird | boat | bottle | bus | car | cat | chair | cow | table | dog | horse | mobike | person | plant | sheep | sofa | train | tv | mAP |
|---|---|---|---|---|---|---|---|---|---|---|---|---|---|---|---|---|---|---|---|---|---|
| PCL Paper[15] | 54.40 | 69.00 | 39.30 | 19.20 | 15.70 | 62.90 | 64.40 | 30.00 | 25.10 | 52.50 | 44.40 | 19.60 | 39.30 | **67.70** | **17.80** | 22.90 | 46.60 | **57.50** | 58.60 | 63.00 | 43.500 |
| PCL PyTorch | 62.30 | 69.35 | 50.59 | 28.10 | 22.07 | **71.80** | 68.10 | 56.81 | 23.91 | 61.29 | 43.13 | 59.40 | 45.02 | 66.21 | 12.32 | 23.28 | 45.32 | 52.05 | 65.09 | 57.18 | 49.170 |
| PCL-Fast + Res50 | 58.90 | 56.19 | 49.12 | 24.35 | 17.07 | 62.42 | 52.25 | 58.98 | 18.38 | 50.14 | 44.95 | 58.03 | 56.27 | 57.53 | 9.27 | 21.23 | 35.22 | 49.00 | 61.31 | 54.06 | 44.739 |
| PCL-Fast + Res101 | 61.34 | 57.35 | 50.37 | 26.37 | 16.44 | 62.30 | 52.35 | 60.46 | 18.16 | 49.40 | 43.42 | 59.32 | 56.96 | 58.50 | 11.15 | 19.80 | 30.89 | 51.16 | 60.01 | 55.52 | 45.063 |
| PCL-Faster + Res50 | **66.60** | 71.18 | 58.88 | **30.32** | 27.14 | 68.18 | **69.17** | 63.21 | **26.78** | **62.51** | **48.79** | 71.82 | **64.60** | 65.61 | 12.52 | 25.43 | **46.73** | 55.98 | **66.26** | 65.07 | **53.163** |
| PCL-Faster + Res101 | 66.19 | **74.12** | **59.40** | 29.19 | **28.04** | 68.82 | 66.17 | **64.98** | 26.51 | 62.43 | 46.77 | **72.54** | 66.32 | 65.29 | 12.41 | **26.64** | 43.85 | 54.86 | 64.62 | **65.11** | **53.212** |

## 5.6 The performance of PGT Generation Strategy

Here, we investigate the influence of different settings of $k$ to generate PGT with Faster RCNN-ResNet101 as the two-phase WSOD architecture. As the curve in Fig.7 shows, the mAP increases steadily from 53.212% to 53.990% as $k$ changes from 1 to 3, although the improvement is slight. The main reason for this improvement is that an image may contain multiple instances from the same categories and a larger $k$ value allows algorithms to consider this case. This confirms the necessity of considering more bounding boxes when generating the PGT. Nonetheless, a different trend can be seen when $k$ continues to increase. The mAP reduces modestly from 53.990% to 53.206% when $k$ increases to 5, and a remarkable reduction (almost 28%) is seen as $k$ changes to 50 as shown in Table 8. These reductions are caused by introducing more negative samples in PGT than positive samples. The results of different values of $k$ for each class on the VOC 2007 test set are listed in Table 8. From Table 8, we can see that the most of the best mAP for categories does not occur in the top one bounding boxes strategy. This confirms our theory.

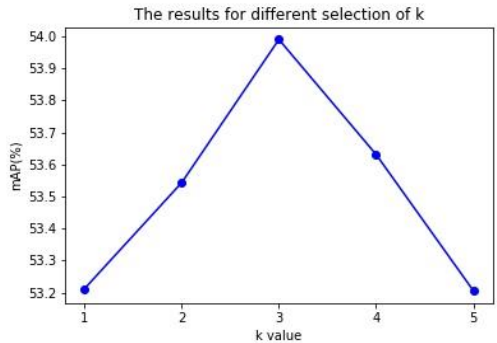

Figure 7. Results (mAP in %) for different $k$ values on VOC 2007 test set.

Table 8. Results (AP in %) for different value of $k$ to create PGT on VOC 2007 test set. The second-phase detector is Faster RCNN and the backbone network is ResNet101.

| $k$ | aero | bike | bird | boat | bottle | bus | car | cat | chair | cow | table | dog | horse | mobike | person | plant | sheep | sofa | train | tv | mAP |
|---|---|---|---|---|---|---|---|---|---|---|---|---|---|---|---|---|---|---|---|---|---|
| 1 | 66.19 | 74.12 | 59.40 | 29.19 | **28.04** | 68.82 | 66.17 | 64.98 | 26.51 | 62.43 | 46.77 | 72.54 | **66.32** | 65.29 | 12.41 | 26.64 | 43.85 | 54.86 | 64.62 | 65.11 | 53.212 |
| 2 | 65.18 | **74.63** | 59.92 | 26.42 | 27.61 | 69.62 | 69.18 | **69.88** | 26.62 | 64.13 | 47.22 | 72.50 | 62.17 | 65.26 | 12.28 | 25.26 | 44.82 | 57.01 | **66.79** | 64.03 | 53.542 |
| 3 | 64.47 | 73.81 | 58.38 | 30.01 | 27.74 | 69.91 | **69.76** | 68.93 | 26.40 | 64.33 | **47.96** | **72.84** | 60.71 | **68.76** | 12.57 | **28.07** | **45.53** | **59.37** | 65.38 | 64.85 | **53.990** |
| 4 | **66.81** | 72.14 | **61.04** | **34.10** | 27.86 | **70.85** | 69.34 | 64.48 | 27.78 | 63.90 | 46.30 | 71.26 | 61.29 | 64.60 | 13.40 | 27.78 | 44.68 | 54.03 | 65.43 | **65.55** | 53.631 |
| 5 | 65.85 | 73.21 | 57.86 | 32.18 | 26.89 | 65.81 | 69.50 | 65.65 | **29.38** | **64.65** | 44.98 | 70.65 | 62.50 | 65.13 | **13.42** | 27.62 | 44.72 | 56.41 | 63.84 | 63.87 | 53.206 |
| 50 | 41.71 | 51.12 | 30.69 | 19.90 | 17.16 | 45.95 | 43.84 | 38.48 | 27.66 | 44.23 | 29.38 | 35.25 | 46.25 | 45.76 | 9.20 | 22.23 | 39.26 | 33.59 | 42.75 | 45.04 | 35.473 |

## 5.7 The performance of PGT Refinement

We also discuss the influence of PGT refinement strategies mentioned in Section 3.3. Experiments are conducted on Faster RCNN-ResNet101 with the top two bounding boxes PGT generation strategy. The reason for selecting this setting has been presented in the experimental setup section. Recall that a complete refining strategy comprises of one strategy from group one and one strategy from group two. We conduct elaborate experiments on all possible combinations of these two groups. As the four curves in Fig. 8 demonstrate, strategy 2 and 3 of group two combining with any strategy from the group one can boost the performance by almost 1% mAP. For these strategies, the algorithm only updates half of the PGT, which prevents the network from significant fluctuation while refining the PGT. Due to the more accurate annotations, the detector is likely to be trained under correct supervision. We can see that the best strategy 1/2 (*/* denotes the $*^{th}$ strategy of the group1/2) improves the mAP from 53.542% to 55.286% (1.74% mAP) which is a competitive and the state-of-the-art performance.

However, strategies {1,2,3}/1 noticeably limit the performance. This reduction is caused by the fluctuation of the network. Strategy 1 of group two updates all the PGT, and the model can not converge if the algorithm frequently performs refinement. A different result can be seen in strategy 4/1 which boosts the mAP by 1%. The reason is that the algorithm performs refinement only once in strategy 4 of group one, and the detector still has adequate training epochs to fit the updated PGT.

Table 9 lists the results for all refinement strategies for each class on the VOC 2007 test set. Strategy 2 of group two updates the higher half of the PGT according to their scores, which means that the second-phase detector believes itself more than the first-phase detector, while strategy 3 of group two is the opposite. By comparing the results of strategy 1/2, 1/3, 2/2, 2/3, we may conclude that it is better to believe the PGT created by PCL if the PGT refinement is performed frequently.

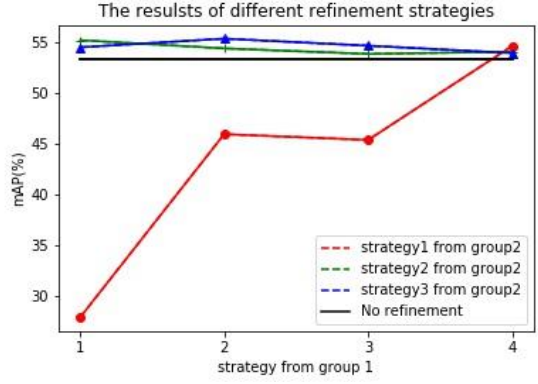

Figure 8. The results of different refinement strategies on VOC 2007 test set. The majority of our strategies boost the mAP by around 1% compared with the model without refinement, while strategy 1 of group two reduces the performance except for combing with strategy four of group 1.

Table 9. The results (AP in %) for different refining strategies for each class on the VOC 2007 test set. */* denotes the $*^{th}$ strategy of the group1/2. -/- means that no refinement is utilized.

| Strategy Group1/2 | aero | bike | bird | boat | bottle | bus | car | cat | chair | cow | table | dog | horse | mobike | person | plant | sheep | sofa | train | tv | mAP |
|---|---|---|---|---|---|---|---|---|---|---|---|---|---|---|---|---|---|---|---|---|---|
| -/- | 65.18 | **74.63** | 59.92 | 26.42 | 27.61 | **69.62** | 69.18 | 69.88 | 26.62 | 64.13 | 47.22 | 72.50 | 62.17 | 65.26 | 12.28 | 25.26 | 44.82 | 57.01 | **66.79** | 64.03 | 53.542 |
| 1/1 | 30.86 | 44.46 | 22.74 | 5.18 | 5.81 | 26.26 | 55.72 | 7.81 | 20.85 | 33.30 | 47.33 | 16.11 | 28.42 | 42.51 | 5.67 | 23.70 | 18.37 | 31.14 | 53.83 | 36.93 | 27.850 |
| 1/2 | 65.03 | 73.21 | **62.13** | 29.13 | 31.33 | 68.04 | 70.83 | 62.35 | **30.23** | 66.01 | **54.03** | 73.79 | 66.38 | 64.81 | **13.96** | 29.13 | 44.12 | 62.08 | 69.45 | 66.67 | 55.135 |
| 1/3 | **68.64** | 73.66 | 60.34 | 24.75 | 31.97 | 67.80 | 67.73 | 66.46 | 28.62 | 59.66 | 52.81 | **74.11** | 65.31 | 65.33 | 13.85 | 26.56 | 46.86 | 60.35 | 67.67 | 66.22 | 54.436 |
| 2/1 | 56.66 | 62.61 | 51.82 | 30.68 | 22.15 | 48.85 | 70.97 | 39.89 | 19.87 | 59.69 | 38.75 | 41.03 | 45.45 | 65.73 | 6.05 | 26.26 | 50.09 | 54.19 | 67.15 | 59.73 | 45.880 |
| 2/2 | 65.85 | 73.21 | 57.86 | **32.18** | 26.89 | 65.81 | 69.50 | 65.65 | 29.38 | 64.65 | 44.98 | 70.65 | 62.50 | 65.13 | 13.42 | 27.62 | 44.72 | 56.41 | 63.84 | 63.87 | 54.325 |
| 2/3 | 65.59 | 72.98 | 61.22 | 29.90 | **34.36** | 69.02 | 67.87 | 60.65 | 28.79 | 67.72 | 52.89 | 71.60 | 68.48 | 67.15 | 12.25 | **30.86** | 43.19 | 59.14 | **72.11** | 69.92 | **55.286** |
| 3/1 | 50.45 | 64.78 | 48.28 | 20.01 | 20.55 | 59.15 | 69.24 | 41.66 | 27.62 | 52.04 | 43.41 | 47.74 | 37.81 | 68.07 | 12.97 | 29.93 | 45.87 | 55.06 | 63.94 | 47.70 | 45.313 |
| 3/2 | 61.93 | 72.27 | 58.47 | 32.13 | 29.86 | 68.45 | 67.43 | **70.21** | 26.49 | 60.17 | 49.45 | 72.34 | 62.23 | 66.07 | 13.35 | 30.59 | 42.56 | **62.81** | 64.52 | 64.30 | 53.782 |
| 3/3 | 66.61 | 74.45 | 60.93 | 25.31 | 30.46 | 69.48 | 70.38 | 65.94 | 28.49 | 60.58 | 50.37 | 73.84 | **68.67** | 66.56 | 12.87 | 26.82 | 47.50 | 60.70 | 68.44 | 63.45 | 54.593 |
| 4/1 | 65.50 | 74.44 | 59.81 | 27.92 | 30.52 | 66.43 | **71.99** | 65.22 | 29.16 | **68.96** | 46.95 | 71.27 | 65.15 | **71.10** | 8.77 | 26.97 | **51.63** | 60.79 | 63.00 | 65.69 | 54.563 |
| 4/2 | 66.28 | 74.22 | 60.65 | 31.77 | 29.02 | 68.30 | 69.99 | 67.31 | 29.17 | 61.52 | 49.59 | 73.25 | 63.05 | 65.77 | 8.26 | 28.24 | 45.83 | 57.63 | 64.28 | 65.87 | 53.999 |
| 4/3 | 64.56 | 72.23 | 59.83 | 26.35 | 28.69 | 67.40 | 66.30 | 66.35 | 29.41 | 64.56 | 48.26 | 72.94 | 65.14 | 66.86 | 12.61 | 27.63 | 44.99 | 59.44 | 66.53 | 67.24 | 53.866 |

## 5.8 Comparison with other methods and Limitations

Finally, we compare our best performed method with other methods. Our best results exploit PCL-Faster-RCNN with Resnet101 as the backbone network and top two PGT generation strategy. In addition, strategy 2/3 is utilized to refine the PGT during the training of Faster RCNN. The results of our best performed method and other methods on the VOC 2007 test set are demonstrated in Table 10. It is obvious that our method outperforms other models. We can expect to obtain similar improvements when applying our methods to other state-of-the-art single WSOD models, which we intend to explore in the future.

However, there are some limitations of our methods. As we stated in Section 3.2, the best-performing value of $k$ of PGT generation strategy may vary depending on the dataset since the distribution of objects in the image is different under different datasets or application scenarios. In addition, refinement strategies slightly increase the computation complexity during the training of the second-phase detector as the algorithm performs inference on the training set and recreates the annotation files in some epochs.

Table 10. Comparison of AP(%) performance on VOC 2007 test set. The last row is our best performed method. FRCNN is the abbreviation of Fast RCNN.

| Method | aero | bike | bird | boat | bottle | bus | car | cat | chair | cow | table | dog | horse | mobike | person | plant | sheep | sofa | train | tv | mAP |
|---|---|---|---|---|---|---|---|---|---|---|---|---|---|---|---|---|---|---|---|---|---|
| PCL-Ens+FRCNN[15] | 63.2 | 69.9 | 47.9 | 22.6 | 27.3 | 71.0 | 69.1 | 49.6 | 12.0 | 60.1 | 51.5 | 37.3 | 63.3 | 63.9 | 15.8 | 23.6 | 48.8 | 55.3 | 61.2 | 62.1 | 48.8 |
| ML-LocNet-L+[20] | 60.8 | 70.6 | 47.8 | 30.2 | 24.8 | 64.9 | 68.4 | 57.9 | 11.0 | 51.3 | 55.5 | 48.1 | 68.7 | 69.5 | **28.3** | 25.2 | 51.3 | 56.5 | 60.0 | 43.3 | 49.7 |
| WSRPN-Ens+FRCNN[21] | 63.0 | 69.7 | 40.8 | 11.6 | 27.7 | 70.5 | **74.1** | 58.5 | 10.0 | 66.7 | **60.6** | 34.7 | **75.7** | **70.3** | 25.7 | 26.5 | 55.4 | 56.4 | 55.5 | 54.9 | 50.4 |
| Multi-Evidence[23] | 64.3 | 68.0 | 56.2 | **36.4** | 23.1 | 68.5 | 67.2 | 64.9 | 7.1 | 54.1 | 47.0 | 57.0 | 69.3 | 65.4 | 20.8 | 23.2 | 50.7 | 59.6 | 65.2 | 57.0 | 51.2 |
| W2F+RPN+FSD2[24] | 63.5 | 70.1 | 50.5 | 31.9 | 14.4 | 72.0 | 67.8 | 73.7 | 23.3 | 53.4 | 49.4 | 65.9 | 57.2 | 67.2 | 27.6 | 23.8 | 51.8 | 58.7 | 64.0 | 62.3 | 52.4 |
| GAM+PCL+Regression[18] | 59.8 | 72.8 | 54.4 | 35.6 | 30.2 | **74.4** | 70.6 | **74.5** | 27.7 | 68.0 | 51.7 | 46.3 | 63.7 | 68.6 | 14.8 | 27.8 | **54.9** | **60.9** | 65.1 | 67.4 | 54.5 |
| Our method | **65.6** | **73.0** | **61.2** | 29.9 | **34.4** | 69.0 | 67.9 | 60.7 | **28.8** | 67.7 | 52.9 | **71.6** | 68.5 | 67.2 | 12.3 | **30.9** | 43.2 | 59.1 | **72.1** | **70.0** | **55.3** |

# 6 CONCLUSION

Instance-level annotation datasets are essential to object detection task. However, it is costly and time-consuming to establish instance-level annotation documents, while image-level annotation is easier, cheaper, and more convenient. Hence, weakly supervised object detection, which trains a detector with image-level annotations, has gained the attention of researchers. In this work, we propose a two-phase WSOD architecture named PCL-Faster RCNN, and some PGT generation strategies to advance the development of WSOD. In addition, we also develop a method to refine the PGT during the training of the second-phase detector.

In two-phase WSOD architectures, the main components are the selection of second-phase detector and the PGT used to train the second-phase detector. We first verify the performance of the two-phase WSOD on the PASCAL VOC 2007 datasets by selecting Fast RCNN and Faster RCNN as second-phase detectors, and top one bounding boxes as PGT. As the results indicate, Faster RCNN achieves a significant 4% improvement compared with the single PCL network and noticeably outperforms the PCL-Fast RCNN architecture. Hence, we propose to select Faster RCNN as the second-phase detector. Moreover, we also investigate the influence of a deeper architecture via conducting experiments on ResNet50 and ResNet101. Experiments demonstrates that there is no noticeable improvement when we replace backbone network ResNet50 with the deeper ResNet101.

Different from traditional methods in this field, we propose to consider more bounding boxes when creating the PGT. As the experimental results indicate, the strategy considering top three bounding boxes improves the mAP to 53.98% (approximately 0.7% increment) on the VOC 2007 test set. Additionally, we develop a method to refine PGT during the training of Faster RCNN. Under the top two PGT generation strategy, our best refinement strategy demonstrates a further 1.74% mAP improvement. Moreover, most of our strategies enhance the mAP by around 1%. The best result integrating all our proposed methods achieves 55.286% mAP which is the state-of-the-art performance.

In the future, we will apply our methods to other state-of-the-art first-phase and second-phase models. In addition, we believe that the hard examples strategy and context information can further improve the performance of our methods.